\documentclass[conference,a4paper]{IEEEtran}
\IEEEoverridecommandlockouts

\usepackage[colorlinks=true, citecolor=blue]{hyperref}
\usepackage[cmex10]{amsmath}
\usepackage{amssymb,amsfonts}
\usepackage{dblfloatfix}

\usepackage[ruled,vlined]{algorithm2e}
\usepackage{graphicx}
\graphicspath{{Figures/PDF/}{Figures/PNG/}}

\usepackage{booktabs}
\usepackage{siunitx}
\usepackage[numbers,compress]{natbib}
\usepackage{texnames}
\usepackage{bm,bbm}
\usepackage{orcidlink}

\usepackage{tabularx}  
\usepackage{array}    
\usepackage{graphicx}
\usepackage{multirow} 
\usepackage{caption}  
\usepackage{subcaption} 
\usepackage{amsmath}

\newcommand{\subparagraph}{} 
\usepackage{titlesec}
\titlespacing{\section}{0pt}{0.2\baselineskip}{0.2\baselineskip}

\begin{document}

\title{Building and Road Recognition in Dense Urban Informal Settlements: A Dataset and Benchmark
\thanks{This work is supported by the Guangdong Provincial Project (No. 2023QN10H717), the Guangzhou-HKUST(GZ) Joint Funding Program (No. 2025A03J3639), and AI Research and Learning Base of Urban Culture Project (No. 2023WZJD008).}
}

\author{	\IEEEauthorblockN{Hongyu Long}
	\IEEEauthorblockA{\textit{UGOD Thrust, HKUST(GZ)}\\
		Guangzhou, China\\
		hlong285@connect.hkust-gz.edu.cn}
	\and
	\IEEEauthorblockN{Jiaxuan Liu\orcidlink{0009-0003-1426-0671}}
	\IEEEauthorblockA{\textit{UGOD Thrust, HKUST(GZ)}\\
		Guangzhou, China\\
		jliu031@connect.hkust-gz.edu.cn}
	\and
	\IEEEauthorblockN{Rui Cao$^*$\orcidlink{0000-0002-1440-4175}}
	\IEEEauthorblockA{\textit{UGOD Thrust, HKUST(GZ)}\\
		Guangzhou, China\\
		ruicao@hkust-gz.edu.cn\\
        *Corresponding author}
}

\maketitle
\begin{abstract}
As a widespread form of informal settlements, urban villages present significant challenges for sustainable urban development and governance. Precise mapping of their infrastructure is essential, however, existing remote sensing datasets primarily focus on formal urban environments, lacking fine-grained annotated data for the high-density building patterns and narrow road networks typical of urban villages. To address this gap, we introduce the \textit{DenseUIS} dataset, the first high-resolution remote sensing dataset specifically designed for building and road extraction in extremely dense urban informal settlements, covering 126 urban villages across Shenzhen and Guangzhou in China. Furthermore, we conduct a comprehensive evaluation of state-of-the-art deep learning models on this dataset. Experimental results reveal the limitations of existing methods in handling the unique morphological patterns of dense informal settlements, underscoring the need for specialized approaches. \textit{DenseUIS} therefore provides a robust benchmark for advancing fine-grained urban mapping in complex and high-density informal environments. The dataset is publicly available at \url{https://github.com/rui-research/DenseUIS}.
\end{abstract}

\begin{IEEEkeywords}
	Informal settlements, urban village, deep learning, semantic segmentation, high-resolution imagery, sustainable development goals (SDGs), GeoAI
\end{IEEEkeywords}

\vspace{1.5em}

\section{Introduction}

\begin{figure}[t] 
    \centering
    \includegraphics[width=\linewidth]{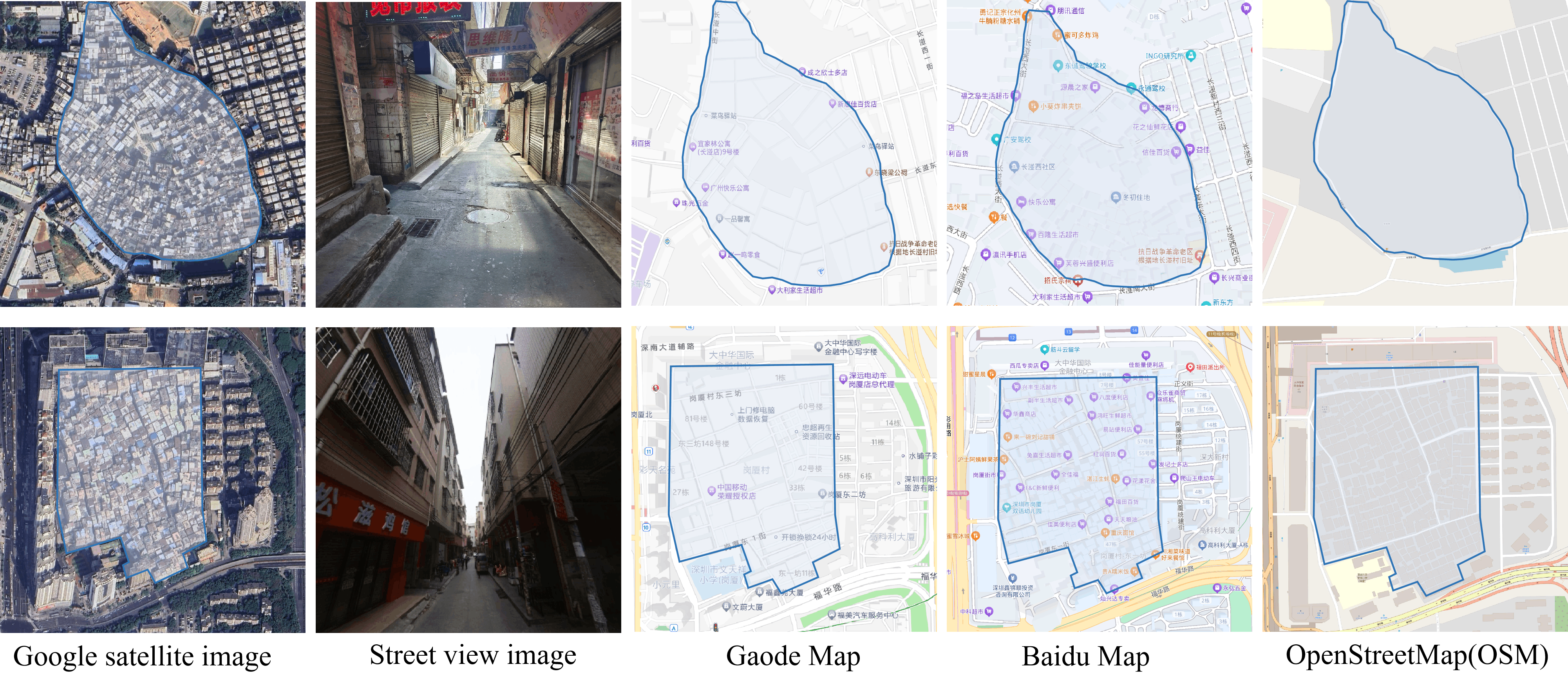}
    \caption{Typical examples of dense urban informal settlements in China, characterized by extremely high building density and intricate road networks. Existing map platforms (e.g., Gaode, Baidu, and OSM) do not provide accurate building and road information for these areas.}
    \label{fig:motivation} 
\end{figure}

Over the past few decades, China has experienced rapid economic growth and urbanization, leading to the emergence of a widespread form of informal settlements known as \textit{urban villages} (UV) \cite{cao2025mapping}. They are typically characterized by narrow and irregular road networks, high building density, limited green space, and inadequate public infrastructure. While these settlements provide affordable housing for rural migrant workers and low-income residents, their substandard living conditions, lack of safety assurances, and high exposure to hazard risks pose serious threats to residents' health and well-being. In line with the United Nations' Sustainable Development Goals (SDGs), improving their living conditions and effectively integrating them into formal urban planning and governance frameworks are a critical societal priority \cite{zhang2025mapping}. Achieving this goal requires precise mapping of fundamental infrastructure, particularly buildings and roads within these areas, which however are difficult to acquire and usually missing in mainstream map platforms, as shown in Fig. \ref{fig:motivation}.

Recent advances in deep learning have significantly improved the extraction of buildings and roads from remote sensing imagery \cite{lu2025deep}. Convolutional Neural Network (CNN)–based architectures such as U-Net \cite{ronneberger2015u} and DeepLab-V3+ \cite{chen2018encoder} effectively integrate multi-scale features through encoder–decoder structures. D-LinkNet \cite{zhou2018d} further enhances contextual understanding by incorporating dilated convolutions to enlarge the receptive field. More recently, Vision Transformers \cite{dosovitskiy2020image} like SegFormer \cite{xie2021segformer} model leverages self-attention mechanisms to capture global spatial dependencies. Given the strong spatial coupling between buildings and roads, multi-task learning frameworks such as JointNet \cite{zhang2019jointnet} and subsequent works \cite{ayala2021deep, abdollahi2021multi}, have demonstrated that jointly extracting both elements leads to substantial performance gains in complex urban environments.

Despite these advances, existing research has predominantly focused on formal urban built-up areas—such as arterial roads, commercial zones, and planned residential districts—while largely overlooking the internal structures of urban villages \cite{lu2025deep}. As illustrated in Fig.~\ref{fig:motivation}, current map platforms, such as Gaode Map \footnote{\url{https://www.gaode.com/}}, Baidu Map \footnote{\url{https://map.baidu.com/}}, and OpenStreetMap (OSM) \footnote{\url{https://www.openstreetmap.org/}}, fail to provide accurate geospatial data of the intricate road networks and densely packed building footprints of UVs. Their narrow alleyways, irregular pathways, and highly compact morphology differ markedly from conventional urban layouts, making navigation difficult even for individuals familiar with conventional city layouts. Addressing this gap requires dedicated datasets that capture the unique morphological and spatial patterns of urban villages. As a core contribution, we construct and introduce the \textit{DenseUIS} dataset of UV buildings and road networks, leveraging high‑resolution remote sensing imagery from hundreds of urban villages in Shenzhen and Guangzhou, which covers diverse urban contexts and structural forms. We then design benchmark experiments to systematically demonstrate the complexity and necessity of this dataset, while validating its practicality for future method development.

The remainder of the paper is organized as follows. Section \ref{sec:Related work} reviews the related research. Section \ref{sec:study area and data} introduces the proposed \textit{DenseUIS} dataset. Section \ref{sec:Experiments} presents the benchmarking experiments and analysis. Finally, Section \ref{sec:conclusion} concludes the paper.

\vspace{1.5em}

\section{Related Work} \label{sec:Related work}

Current studies on urban village mapping primarily focus on delineating their spatial boundaries, yet often overlook the critical internal structures such as buildings and roads \cite{cao2025mapping,zhang2025mapping,zhou2025semi}. Recent advancements in UV analysis have evolved from multi-modal classification to large-scale benchmarking. Early works, such as S$^2$UV~\cite{chen2022multi} and RsSt-ShenzhenUV~\cite{fan2022multilevel}, integrated satellite and street-view imagery to address spectral confusion in UV identification. Subsequently, datasets like LtCUV~\cite{lin2024long} and CUGUV~\cite{wang2025cuguv} expanded the scope to nationwide and long-term monitoring, facilitating cross-city generalization studies. Additionally, UV-SAM~\cite{zhang2024uvsam} adapted the Segment Anything Model~\cite{kirillov2023segment} for precise boundary extraction via prompt learning. However, these existing datasets primarily focus on determining the external boundaries of UVs, leaving the fine-grained coupling of internal building footprints and road networks largely unexplored.

Currently, there are numerous public road datasets to support road extraction tasks, such as the Massachusetts Roads Dataset~\cite{mnih2013machine}, the DeepGlobe Roads Extraction Dataset~\cite{demir2018deepglobe}, the SpaceNet Road Dataset~\cite{vanetten2018spacenet}, and the CHN6-CUG Road Dataset~\cite{zhu2021global}. Although these datasets have significantly advanced the development of deep learning-based road extraction methods, they focus more on urban arterial roads, leaving a notable gap in the extraction of internal road networks within urban villages. Due to the constrained internal spaces and complex backgrounds characteristic of urban villages, these roads are typically narrower and more challenging to identify compared to standard urban roads, particularly in areas with high building density. While the WHU-RUR+ Rural Road Dataset~\cite{wang2025whu} effectively addresses road identification in rural non-urban core areas, instances of building enclosure are relatively scarce in rural settings; consequently, this dataset does not account for the mutual coupling between buildings and roads. 

Existing large-scale building datasets have established a strong foundation for urban mapping. Prominent works such as CBRA~\cite{liu2023cbra}, CBF~\cite{huang2024china}, and GABLE~\cite{sun2024gable} have successfully mapped building footprints at the national level, while others like 3D-GloBFP~\cite{che20243dglobfp} and the East Asia dataset~\cite{shi2024last} extend this coverage to regional or global scales. Although these products provide valuable macro-level insights, they generally lack the granular detail necessary to accurately represent complex informal settlements. 

To address these limitations, we propose the \textit{DenseUIS} dataset, which is the first high-resolution building and road network dataset specifically constructed for urban village scenarios.
Our dataset stands out as a unique contribution dedicated specifically to dense urban villages. Unlike previous general-purpose collections, ours fills a critical gap by capturing the intricate morphology of these densely packed environments, offering an exclusive, high-precision resource for fine-grained urban studies.

\vspace{1.5em}

\section{Overview of the \textit{DenseUIS} Dataset} \label{sec:study area and data}
\subsection{Study area}
\begin{figure}[htbp] 
    \centering
    \includegraphics[width=\linewidth]{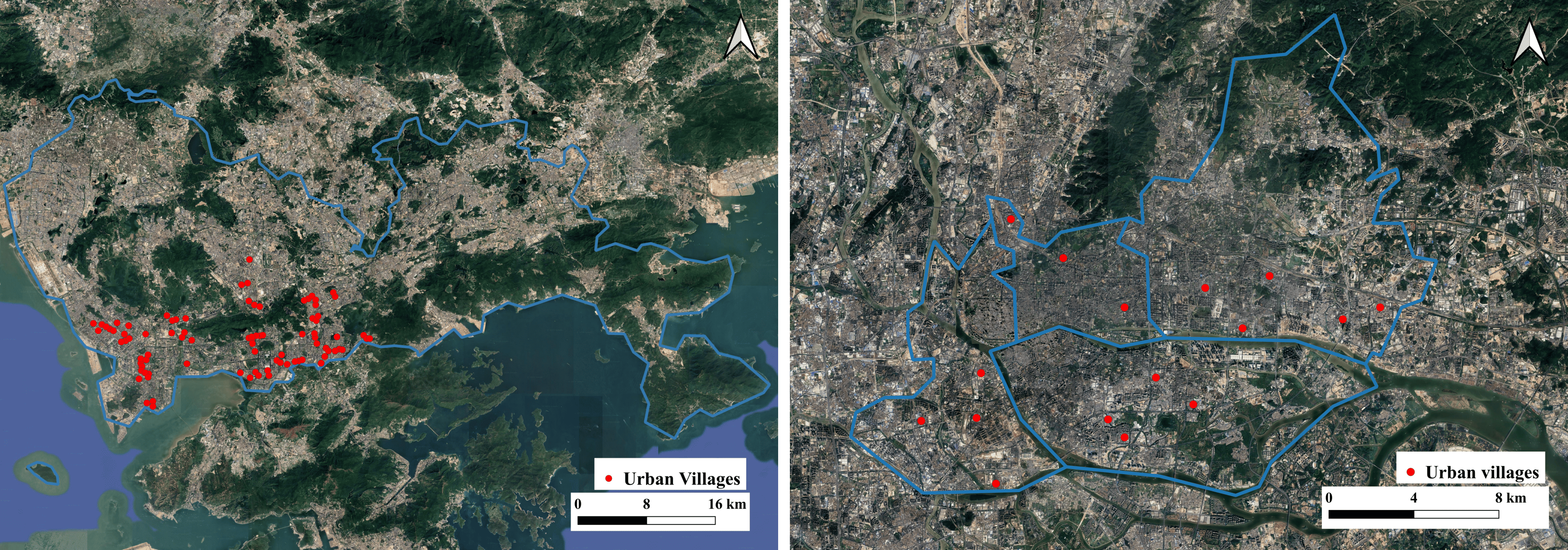}
    \caption{Geographic distribution of \textit{DenseUIS} dataset samples. The two regions in the map represent Shenzhen (left) and Guangzhou (right), respectively.}
    \label{fig:image2} 
\end{figure}

The \textit{DenseUIS} dataset is designed to benchmark building and road extraction in complex urban environments, specifically targeting dense urban villages. To ensure model generalizability, the dataset encompasses substantial intra-class diversity in terms of building density, scale, and spatial layout. Specifically, we focus on urban villages in Shenzhen and Guangzhou, China—regions characterized by very high building density and irregular, narrow road networks.
The dataset covers 126 urban villages, sampled from Futian, Luohu, Nanshan, Bao’an, Longhua, and Longgang districts in Shenzhen, as well as from the central districts of Guangzhou, including Yuexiu, Liwan, Haizhu, and Tianhe.
The spatial distribution of the dataset samples is illustrated in Fig.~\ref{fig:image2}.

\subsection{Annotation method}
We utilized the high-resolution Google Earth imagery at the zoom level of 20 (with spatial resolution of appropriately $0.14 m$) as the primary source for manual annotation, with existing building and road data from Gaode and Tianditu served as auxiliary references.
The annotation comprises four main steps:
\begin{itemize}
    \item Step 1: The building data, road network data,  and high-resolution remote sensing imagery were projected onto the same coordinate system to maintain geographical consistency. 
    \item Step 2: In urban village areas, inaccurate, missing, or unnecessary road centerlines and buildings were manually corrected or removed. Concurrently, road networks were manually annotated by inferring passable routes through gaps between buildings.
    \item Step 3: We generated pixel-level annotations for both road networks and buildings by establishing buffer zones along their refined vector boundaries. To accommodate regional differences, distinct buffer widths were applied to Shenzhen and Guangzhou, respectively. The resulting roads and buildings buffered areas were defined as positive regions, while all remaining pixels were categorized as background.
    \item Step 4: The annotated data of road networks and buildings, along with the high-resolution remote sensing imagery, were cropped into $1024 \times 1024$ patches. 
\end{itemize}

\subsection{Characteristics of the DenseUIS dataset}

The \textit{DenseUIS} dataset is constructed from 1,000 very-high-resolution satellite images. 
The dataset covers two typical urban village morphologies: disorderly arranged layouts and more regularly arranged settlements. A key annotation feature is that many labeled “roads” correspond to visible gaps between buildings, rather than clear paved surfaces, due to frequent occlusion and the narrow, irregular nature of pathways in these dense areas. Although these annotations do not always represent formal roads, they provide a challenging benchmark that tests a model’s ability to infer navigable space and segment complex urban textures under highly ambiguous conditions, thus rigorously evaluating segmentation and reasoning capabilities in informal settlement contexts.

\begin{figure}[t] 
    \centering
    
    \includegraphics[width=\linewidth]{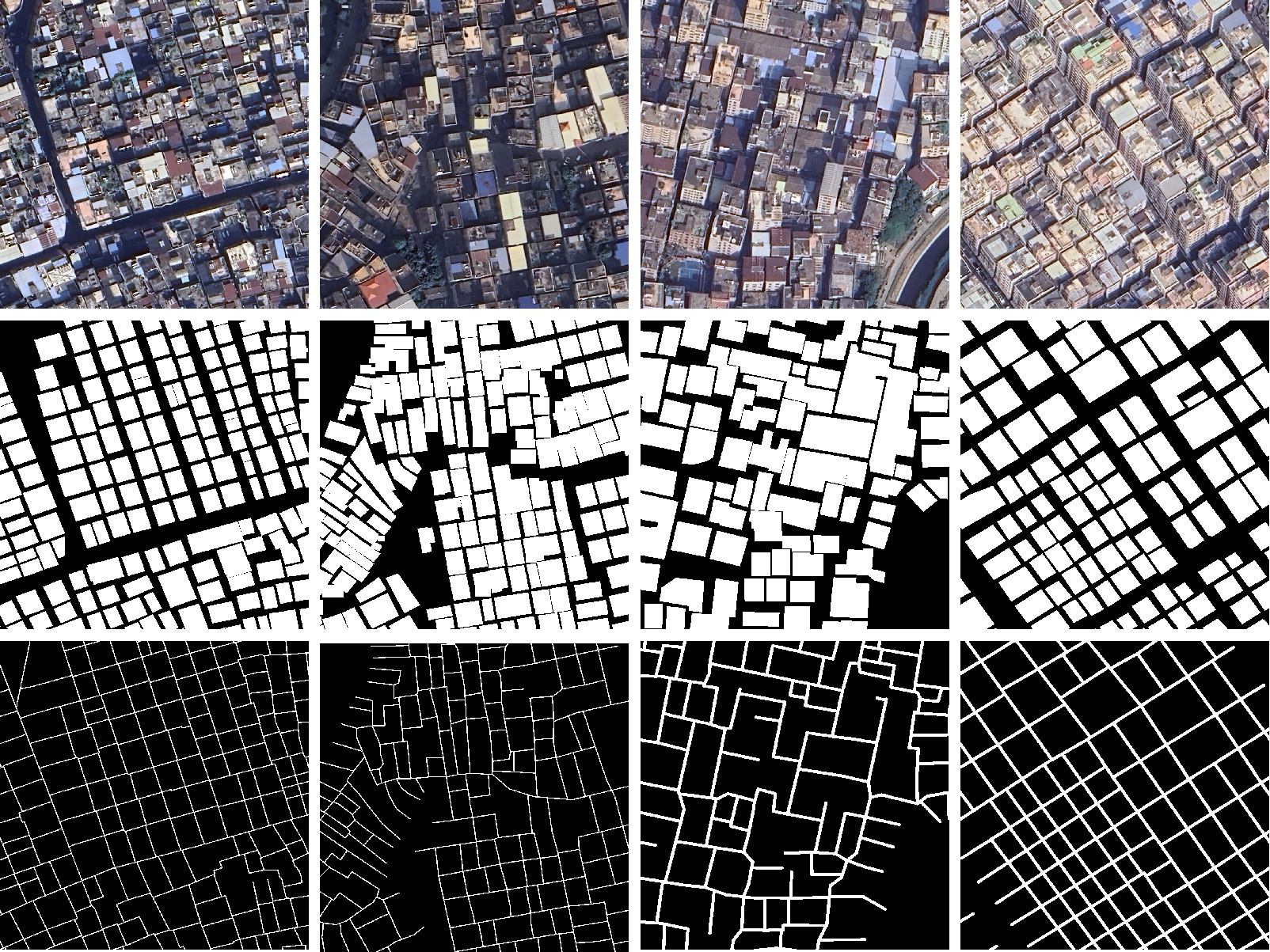}
    
    \caption{Examples of the buildings and roads in the \textit{DenseUIS} dataset.}
    \label{fig:image3} 
\end{figure}

Compared to existing building and road datasets, \textit{DenseUIS} is distinguished by its unique focus on informal settlements, higher spatial resolution, and significantly greater scene complexity. Fig. \ref{fig:image3} presents typical examples of the \textit{DenseUIS} dataset. Key features are summarized as follows:

\textit{1) Unique study area:} \textit{DenseUIS} focuses on a distinctive settlement type in China’s urbanization process: the Urban Village. The road data within this context includes diverse typologies specific to such environments, such as renovated village roads, unrenovated narrow alleys, and connecting arterial roads. The building data feature small roof footprints and high spatial density.

\textit{2) High spatial resolution and large image size:} Higher-resolution remote sensing imagery enables finer extraction of urban features. In dense urban villages with narrow pathways, such detail is essential for accurate road mapping. The dataset offers high spatial resolution, supporting precise feature delineation, while its large image size aids in modeling building–road relationships to improve model generalization.

\textit{3) Extremely complex scenes:} The dataset highlights the distinct contrast between the dense, irregular layout of urban villages and surrounding formal urban areas, helping models identify intricate roads and dense buildings within such settlements. Annotations are based on the principle that passable routes exist between most buildings, even where road surfaces are visually indistinct.

\vspace{1.5em}

\section{Benchmarking Experiments and Analysis} \label{sec:Experiments}

\subsection{Implementation details}
We conducted all experiments on a server running Ubuntu 22.04 LTS, equipped with an Intel Xeon Platinum 8358 CPU and an NVIDIA GeForce RTX 4090 GPU. Models were implemented in PyTorch and trained using a consistent setup: input images were cropped to $1024\times 1024$ pixels and augmented via random flipping, transposition, and rotation. The AdamW optimizer was employed with a batch size of 4 for 100 epochs, using a momentum of 0.9 and a weight decay of 0.00005. The dataset was divided into 700 samples for training, 100 for validation, and 200 for testing.
\subsection{Quantitative experiment results}
To objectively evaluate the performance of different methods on the proposed dataset, we designed several benchmark experiments and calculated four standard metrics for comparison: Precision, Recall, F1-Score, and Intersection over Union (IoU). The quantitative results of building and road extraction are summarized in Table \ref{tab:building_sub} and Table \ref{tab:road_sub}, respectively.

In building extraction task, U-Net~\cite{ronneberger2015u} sets the baseline with $67.03\%$ IoU. DeepLab-V3+~\cite{chen2018encoder} significantly improves the performance to $72.39\%$ by leveraging multi-scale context features. SegFormer~\cite{xie2021segformer} further elevates the IoU to $75.28\%$ and attains the highest Precision ($88.35\%$) via hierarchical transformers. Ultimately, RS-Mamba~\cite{zhao2024rsmamba} achieves the state-of-the-art with the highest F1-Score ($86.36\%$) and IoU ($75.99\%$). Notably, its superior Recall ($87.22\%$) demonstrates that the Mamba architecture excels at minimizing missed detections in dense building environments compared to CNN and Transformer baselines. 

In road extraction task, U-Net~\cite{ronneberger2015u} achieves high Precision ($65.90\%$) but suffers from low Recall ($39.74\%$). D-LinkNet~\cite{zhou2018d} improves Recall to $54.23\%$ via dilated convolutions, while HRNet~\cite{wang2020deep} boosts the F1-Score to $60.50\%$ by preserving high-resolution details. Notably, RS-Mamba~\cite{zhao2024rsmamba} outperforms all CNN baselines, attaining the highest Recall ($65.36\%$) and IoU ($45.92\%$). This validates the superiority of Mamba's global modeling over local convolutions for extracting continuous road networks.

\begin{table}
    \centering
    \caption{Quantitative results of building extraction.}
    \begin{tabular}{lcccc}
        \toprule
        \textbf{Methods} & \textbf{Precision (\%)} & \textbf{Recall (\%)} & \textbf{F1 (\%)} & \textbf{IoU (\%)} \\
        \midrule
        U-Net       & 81.89 & 78.70 & 80.26 & 67.03 \\
        DeepLab-V3+  & 86.56 & 81.56 & 83.98 & 72.39 \\
        SegFormer   & \textbf{88.35} & 83.58 & 85.90 & 75.28 \\
        RS-Mamba & 85.51 & \textbf{87.22} & \textbf{86.36} & \textbf{75.99} \\
        \bottomrule
    \end{tabular}
    \label{tab:building_sub}
\end{table}

\begin{table}
    \centering
    \caption{Quantitative results of road extraction.}
    \begin{tabular}{lcccc}
        \toprule
        \textbf{Methods} & \textbf{Precision (\%)} & \textbf{Recall (\%)} & \textbf{F1 (\%)} & \textbf{IoU (\%)} \\
        \midrule
        U-Net       & \textbf{65.90} & 39.74 & 49.58 & 32.96 \\
        D-LinkNet  & 64.03 & 54.23 & 58.73 & 41.57 \\
        HRNet   & 62.62 & 58.51 & 60.50 & 43.37 \\
        RS-Mamba & 60.83 & \textbf{65.36} & \textbf{62.94} & \textbf{45.92} \\
        \bottomrule
    \end{tabular}
    \label{tab:road_sub}
\end{table}

\subsection{Qualitative experiment results}

\begin{figure}[t] 
    \centering
    \renewcommand{\thesubfigure}{\Alph{subfigure}}

    \begin{subfigure}[b]{\linewidth}
        \centering
        \includegraphics[width=\linewidth]{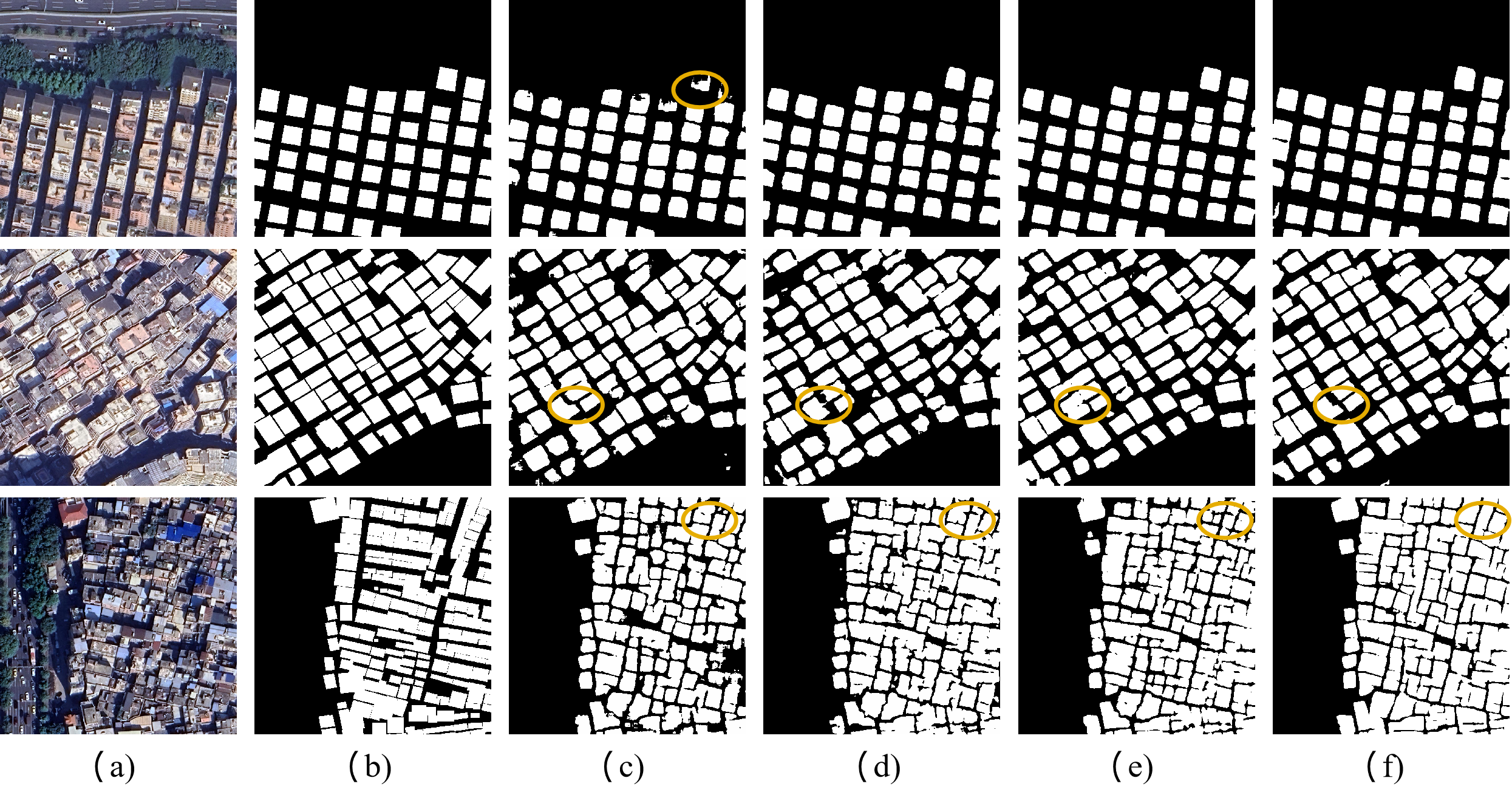}
        \caption{Building extraction results}
        \label{fig:buildings}
    \end{subfigure}

    \vspace{0.2cm} 

    \begin{subfigure}[b]{\linewidth}
        \centering
        \includegraphics[width=\linewidth]{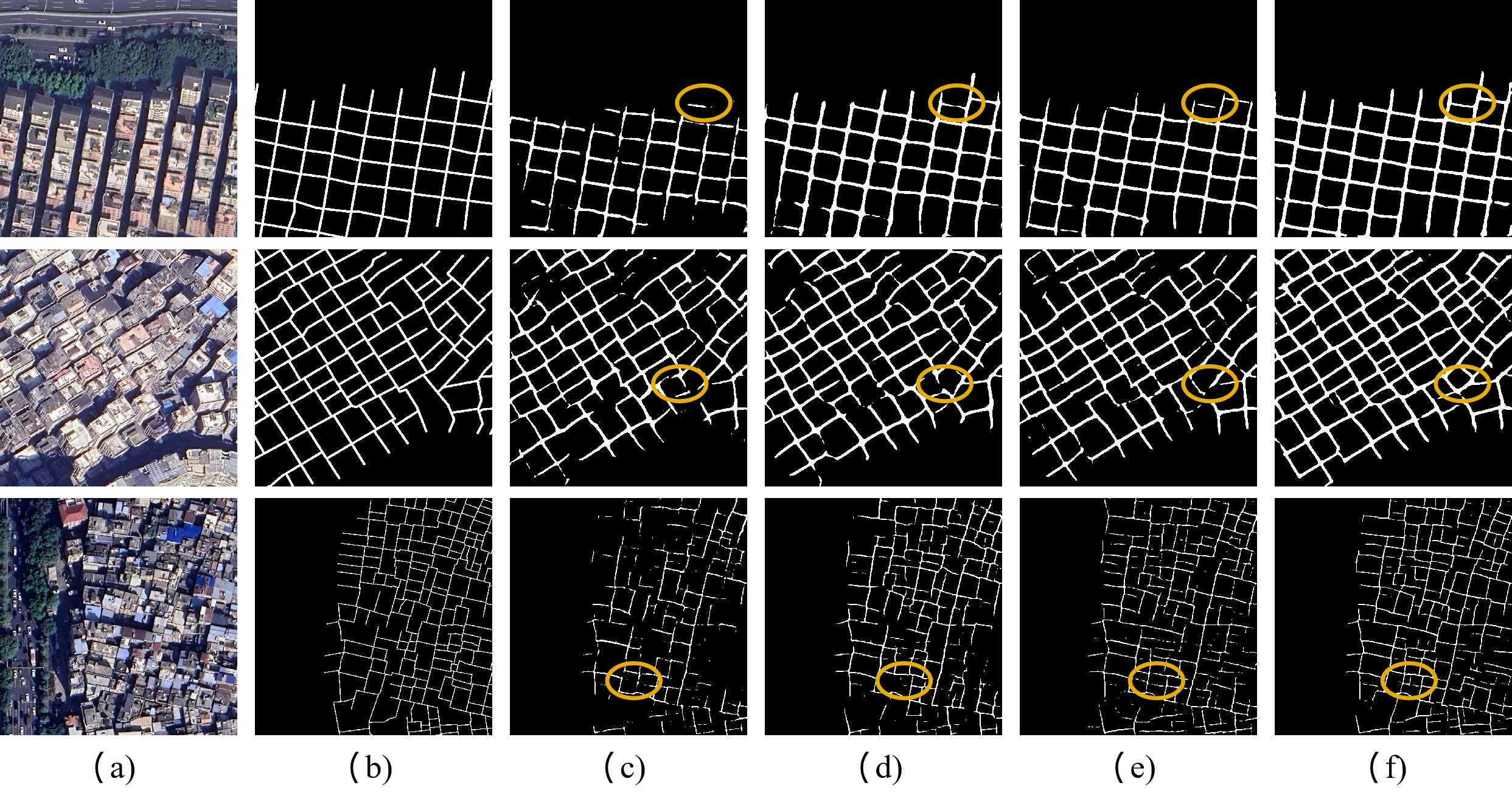} 
        \caption{Road extraction results} 
        \label{fig:roads} 
    \end{subfigure}
    
    \caption{Visual comparison of the prediction results. For building extraction task: (a) Satellite image, (b) Ground truth, (c) U-Net, (d) DeepLab-V3+, (e) SegFormer, (f) RS-Mamba. For road extraction task: (a) Satellite image, (b) Ground truth, (c) U-Net, (d) HRNet, (e) D-LinkNet, (f) RS-Mamba.}
    \label{fig:visual_results}
\end{figure}

Fig. \ref{fig:visual_results} presents exemplar results qualitatively.
Specifically, Fig. \ref{fig:buildings} compares building extraction results across different methods on the test set. CNN-based methods (U-Net, DeepLab-V3+) often struggle to delineate the boundaries of densely packed buildings, as shown in columns (c) and (d). They exhibit the adhesion effect, where adjacent buildings are incorrectly merged into a single object, and they frequently miss small-scale structures, as highlighted by the yellow circles. While Transformer-based method (SegFormer) improves the internal consistency of building footprints, it still produces some irregular boundaries in complex scenes. In contrast, RS-Mamba (f) demonstrates better performance in boundary preservation. It effectively captures global context to separate adjacent instances and generates sharp, regular geometric shapes that are consistent with the Ground Truth.

Similarly, Fig. \ref{fig:roads} compares road extraction results. As observed in columns (c) and (d), the CNN-based methods, U-Net and HRNet, suffer from limited receptive fields, leading to severe fragmentation. The predicted road networks are often discontinuous, with numerous broken segments and missing intersections (indicated by the yellow markers). While specialized method, D-LinkNet (e), improves connectivity by utilizing dilated convolutions, it still fails to recover fine details in occluded areas. Conversely, RS-Mamba (f) achieves the most robust performance. Leveraging its ability to model long-range dependencies, it successfully reconstructs complete road networks with high topological integrity, effectively bridging the gaps that other methods fail to detect.

\section{Conclusion} \label{sec:conclusion}
In this paper, we propose the \textit{DenseUIS} dataset, a novel remote sensing dataset designed for building and road recognition in highly dense urban informal settlements. Spanning 126 urban villages across Shenzhen and Guangzhou, China, this dataset provides very-high-resolution imagery with detailed semantic annotations that capture the intricate spatial patterns of internal road networks and densely packed buildings. It exhibits high spatial resolution, significant heterogeneity, and complex background conditions. By benchmarking widely used deep learning methods, we establish a baseline for extraction tasks and demonstrate both the comprehensiveness and the inherent difficulty of the dataset. Future work will explore more effective deep learning approaches for building and road extraction within dense informal environments. We anticipate that \textit{DenseUIS} will not only support the development of advanced and robust deep learning methods for challenging urban scenarios, but also enable finer‑grained, human‑centric urban analysis and governance, ultimately contributing to global sustainable development goals.

\vspace{0.5em}

\small
\bibliographystyle{IEEEtranN}
\bibliography{references}

\end{document}